%% file: main.tex
\documentclass[runningheads]{llncs}
\usepackage{graphicx}
\newcommand{\citep}{\cite}
\newcommand{\citet}{\cite}
\newcommand\bycite[1]{in~\citet{#1}}

\input{config}

\input{preamble}

\begin{document}
 
\input{metadata.tex}

\maketitle              
 
\begin{abstract}
  \input{abstract}
\end{abstract}

\section{Introduction}
\label{sec:introduction}
\input{intro}

\section{Proposed method}
\label{sec:proposed-method}
\input{method}

\section{Experiments}
\label{sec:experiments}
\input{experiments}

\section{Results}
\label{sec:results}
\input{results}

\section{Hyperparameters and ablation studies}
\label{sec:ablation-studies}
\input{ablation}

\section{Conclusions and further research}
\label{sec:conclusions}
\input{conclusions}

\bibliographystyle{splncs04}
\bibliography{bibliography}

\end{document}

%% file: config.tex
\newif\ifwithappendix
\withappendixfalse

\newif\ifappendixshown
\appendixshowntrue

%% file: preamble.tex
\usepackage{amssymb,amsmath}
\usepackage{caption,subcaption}
\usepackage{multirow}
\usepackage{placeins}
\usepackage{colortbl}
\usepackage{dcolumn}

\usepackage{etoolbox,siunitx}
\robustify\bf
\robustify\bfseries
\robustify\textbf
\robustify\textit
\newcommand{\RR}{\mathbb{R}}

\newcommand{\gray}{\color{black!33}}

\usepackage[T1]{fontenc}
\usepackage[utf8]{inputenc}
\usepackage[textsize=tiny]{todonotes}
\usepackage{graphicx}
\usepackage{booktabs}
\usepackage{latexsym}

\usepackage{footnote}
\makesavenoteenv{tabular}
\makesavenoteenv{table}
\makesavenoteenv{table*}

\usepackage{xurl}

\newcommand\minput[1]{%
  \input{#1}%
  \ifhmode\ifnum\lastnodetype=11 \unskip\fi\fi}

\usepackage{hyperref}
\usepackage{xstring}

\newif\ifaclfinal
\aclfinalfalse

\newcommand{\noqa}[1]{}
\newcommand{\noqall}[1]{}


%% file: metadata.tex
\titlerunning{LAMBERT: Layout-Aware Language Modeling}

\title{LAMBERT: Layout-Aware Language Modeling for Information Extraction}

\author{
  Łukasz Garncarek\inst{1}\thanks{corresponding author; the first four authors
    have equally contributed to the paper.} \and
  Rafał Powalski\inst{1} \and
  Tomasz Stanisławek\inst{1,2} \and
  Bartosz~Topolski\inst{1} \and
  Piotr Halama\inst{1} \and
  Michał Turski\inst{1,3} \and
  Filip~Graliński\inst{1,3} 
}

\authorrunning{Ł. Garncarek et al.}

\institute{
    Applica.ai, Zajęcza 15, 00-351 Warszawa, Poland\\
    \email{firstname.lastname@applica.ai}
    \and
    Warsaw University of Technology, Koszykowa 75, 00-662 Warszawa, Poland
     \email{firstname.lastname@pw.edu.pl}
    \and
    Adam Mickiewicz University, 1 Wieniawskiego, 61-712 Poznań, Poland
     \email{firstname.lastname@amu.edu.pl}
  }


%% file: abstract.tex
We introduce a simple new approach to the problem of understanding documents
where non-trivial layout influences the local semantics. To this end, we modify the
Transformer encoder architecture in a way that allows it to use layout features
obtained from an OCR system, without the need to re-learn language semantics
from scratch. We only augment the input of the model with the coordinates of
token bounding boxes, avoiding, in this way, the use of raw images. This leads to a
layout-aware language model which can then be fine-tuned on downstream tasks.

The model is evaluated on an end-to-end information extraction task using four
publicly available datasets: Kleister NDA, Kleister Charity, SROIE and CORD. We
show that our model achieves superior performance on datasets consisting of
visually rich documents, while also outperforming the baseline RoBERTa on
documents with flat layout (NDA \(F_{1}\) increase from 78.50 to 80.42). Our
solution ranked first on the public leaderboard for the Key Information
Extraction from the SROIE dataset, improving the SOTA \(F_{1}\)-score from 97.81
to 98.17.

\keywords{Language model \and Layout \and Key information extraction \and
  Transformer \and Visually rich document \and Document understanding}


%% file: intro.tex
The sequential structure of text leads to it being treated as a~sequence of
tokens, characters, or more recently, subword units. In many problems related to
Natural Language Processing (NLP), this linear perspective was enough to enable
significant breakthroughs, such as the introduction of the neural Transformer
architecture~\citep{Vaswani2017-transformer}. In this setting, the task of
computing token embeddings is solved by Transformer encoders, such as
BERT~\citep{devlin2019-bert} and its derivatives, achieving top scores on the
GLUE benchmark~\citep{wang2019-glue}.

They all deal with problems arising in texts defined as sequences of words.
However, in many cases there is a structure more intricate than just a linear
ordering of tokens. Take, for instance, printed or richly-formatted documents,
where the relative positions of tokens contained in tables, spacing between
paragraphs, or different styles of headers, all carry useful information. After
all, the goal of endowing texts with layout and formatting is to improve
readability.

In this article we present one of the first attempts to enrich the
state-of-the-art methods of NLP with layout understanding mechanisms,
contemporaneous with \citep{xu2019layoutlm}, to which we compare our model. Our
approach injects the layout information into a pretrained instance
of RoBERTa. We fine-tune the augmented model on a dataset consisting
of documents with non-trivial layout.

We evaluate our model on the end-to-end information extraction task, where the
training set consists of documents and the target values of the properties to be
extracted, without any additional annotations specifying the locations where the
information on these properties can be found in the documents. We compare the
results with a baseline RoBERTa model, which relies on the sequential order of
tokens obtained from the OCR alone (and does not use the layout features), and with
the solution of \citet{xu2019layoutlm,xu2020layoutlmv2}. LAMBERT achieves
superior performance on visually rich documents, without sacrificing results on
more linear texts.

\subsection{Related work}
\label{sec:related-work}

There are two main lines of research into understanding documents with
non-trivial layout. The first one is Document Layout Analysis
\noqa{spell-DLA}(DLA), the goal of which is to identify contiguous blocks of
text and other non-textual objects on the page and determine their function and
order in the document. The obtained segmentation can be combined with the
textual information contained in the detected blocks. This kind of method has
recently been employed \bycite{Liu2019-alibaba}.

Many services employ \noqa{spell-DLA}DLA functionality for OCR (which requires
document segmentation), table detection or form field detection, and their
capabilities are still expanding. The most notable examples are Amazon
Textract~\citep{amazon-textract}, the Google Cloud Document Understanding AI
platform~\citep{google-cloud}, and Microsoft Cognitive
Services~\citep{microsoft-cognitive}. However, each has limitations, such as the
need to create rules for extracting information from the tables recognized by
the system, or use training datasets with annotated document segments. More
recent works on information extraction using DLA include, among others,
\citet{Ishitani2001,Cesarini2003,Hamza2008,bart2010,Medvet2011,Peanho2012,Rusinol2013}.
They concentrate on specific types of documents, such as invoices or forms,
where the layout plays a relatively greater role: more general documents may
contain tables, but they can also have large amounts of unstructured text.

The second idea is to directly combine the methods of Computer Vision and NLP.
This could be done, for instance, by representing a text-filled page as a
multi-channel image, with channels corresponding to the features encoding the
semantics of the underlying text, and, subsequently, using convolutional
networks. This method was used, among others, by Chargrid and BERTgrid models
\cite{Katti2018-chargrid,denk2019-bertgrid}. On the other hand, LayoutLM
\citep{xu2019layoutlm} and TRIE \cite{Zhang2020TRIE} used the image recognition
features of the page image itself. A more complex approach was taken by PICK
\cite{Yu2020PICK}, which separately processes the text and images of blocks
identified in the document. In this way it computes the vertex embeddings of the
block graph, which is then processed with a graph neural network.

Our idea is also related to the one used \bycite{rahman2019-mbert}, though in a
different setting. They considered texts accompanied by audio-visual
signal injected into a pretrained BERT instance, by combining it
with the input embeddings.

LAMBERT has a different approach. It uses neither the raw document image, nor
the block structure that has to be somehow inferred. It relies on the tokens and
their bounding boxes alone, both of which are easily obtainable from any
reasonable OCR system.

\subsection{Contribution}
\label{sec:contribution}

Our main contribution is the introduction of a \emph{Layout-Aware Language
  Model}, a general-purpose language model that views text not simply as a
sequence of words, but as a collection of tokens on a two-dimensional page. As
such it is able to process plain text documents, but also tables, headers, forms
and various other visual elements. The implementation of the model is available
at \url{https://github.com/applicaai/lambert}.

A key feature of this solution is that it retains the crucial trait of language
models: the ability to learn in an unsupervised setting. This allows the
exploitation of abundantly available unannotated public documents, and a
transfer of the learned representations to downstream tasks. Another advantage
is the simplicity of this approach, which requires only an augmentation of the
input with token bounding boxes. In particular, no images are needed. This
eliminates an important performance factor in industrial systems, where large
volumes of documents have to be sent over a network between distributed
processing services.

Another contribution of the paper is an extensive ablation study of the impact
of augmenting RoBERTa with various types of additional positional embeddings on
model performance on the SROIE \citep{icdar-sroie}, CORD \citep{park2019cord},
Kleister NDA and Kleister Charity datasets \citep{graliski2020kleister}.

Finally, we created a new dataset for the unsupervised training of layout-aware
language models. We will share a 200k document subset, amounting to 2M visually
rich pages, accompanied by a dual classification of documents: business/legal
documents with complex structure; and others. Due to IIT-CDIP Test Collection
dataset \cite{Lewis2006} accessibility problems\footnote{the link
  \url{https://ir.nist.gov/cdip/} seems to be dead (access on Feb 17, 2021)},
this would constitute the largest widely available dataset for training
layout-aware language models. It would allow researchers to compare the
performance of their solutions not only on the same test sets, but also with the
same training set. The dataset is published at
\url{https://github.com/applicaai/lambert}, together with a more detailed description
that is too long for this paper.


%% file: method.tex
\begin{figure}[htb]
  \centering
  \includegraphics[width=122mm]{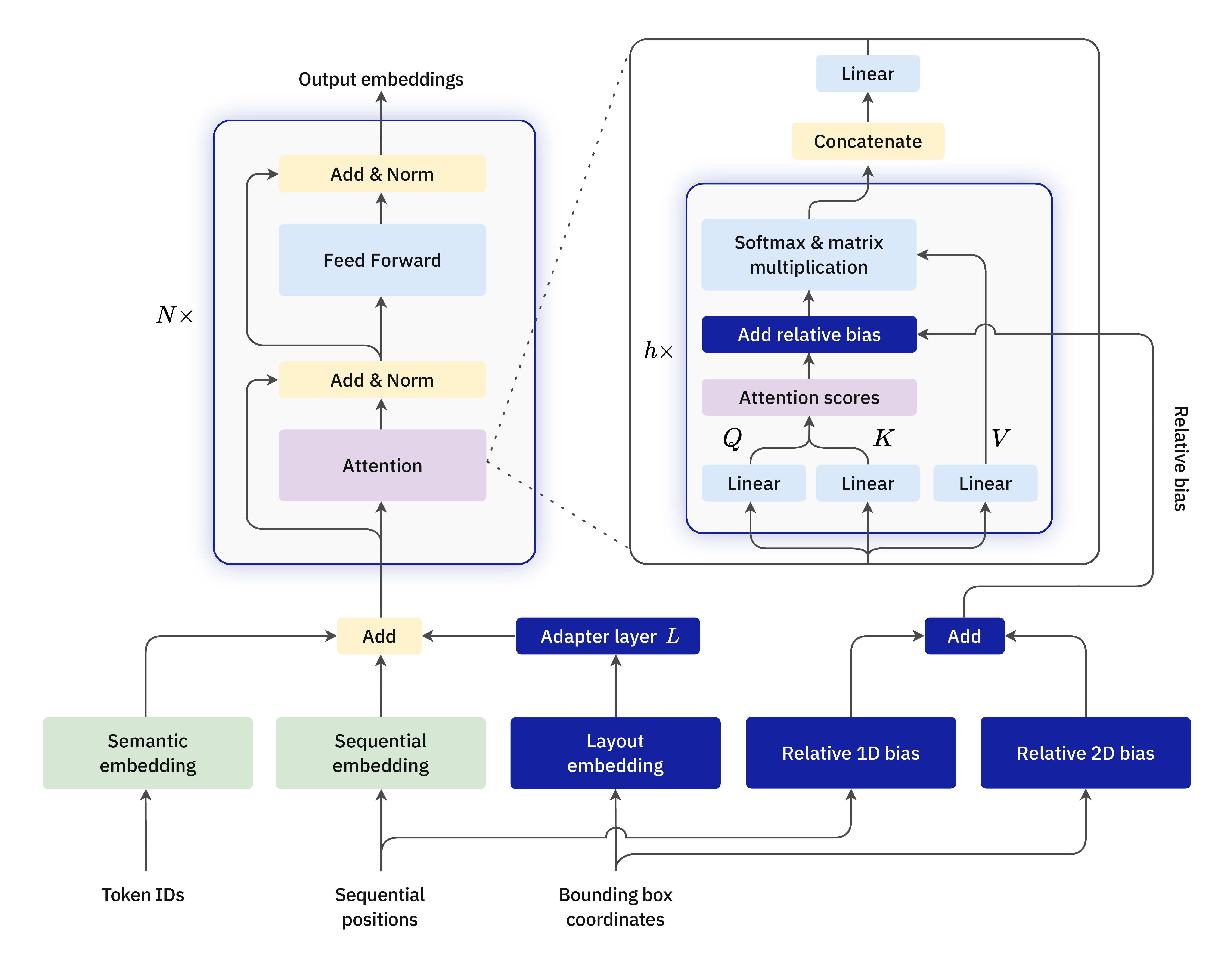}
  \caption{LAMBERT model architecture. Differences with the plain RoBERTa model
    are indicated by white text on dark blue background. \(N=12\) is the number
    of transformer encoder layers, and \(h=12\) is the number of attention heads
    in each encoder layer. \(Q\), \(K\), and \(V\) are, respectively, the queries,
    keys and values obtained by projecting the self-attention inputs.}
  \label{fig:architecture}
\end{figure}

We inject the layout information into the model in two ways. Firstly, we modify
the input embeddings of the original RoBERTa model by adding the layout term. We
also experiment with completely removing the sequential embedding term.
Secondly, we apply relative attention bias, used
\bycite{shaw2018,Huang2019,Raffel2020} in the context of sequential position.
The final architecture is depicted in Figure~\ref{fig:architecture}.

\subsection{Background}
\label{sec:bert}

The basic Transformer encoder, used in, for instance, BERT
\citep{devlin2019-bert} and RoBERTa \citep{liu2019-roberta}, is a
sequence-to-sequence model transforming a sequence of input embeddings
\(x_i\in\RR^n\) into a sequence of output embeddings \(y_i\in\RR^m\) of the same
length, for the input/output dimensions \(n\) and \(m\). One of the main
distinctive features of this architecture is that it discards the order of its
input vectors. This allows parallelization levels unattainable for recurrent neural
networks.

In such a setting, the information about the order of tokens is preserved not by
the structure of the input. Instead, it is explicitly passed to the model, by
defining the input embeddings as
\begin{equation}
  \label{eq:bert-emb}
  x_i = s_i + p_i,
\end{equation}
where \(s_i\in\RR^n\) is the semantic embedding of the token at position \(i\),
taken from a trainable embedding layer, while \(p_i\in\RR^n\) is a
\emph{positional embedding}, depending only on \(i\). In order to avoid
confusion, we will, henceforth, use the term \emph{sequential embeddings}
instead of \emph{positional embeddings}, as the \emph{positional} might be
understood as relating to the 2-dimensional position on the page, which we will
deal with separately.

Since in RoBERTa, on which we base our approach, the embeddings \(p_i\) are
trainable, the number of pretrained embeddings (in this case~512) defines a
limit on the length of the input sequence. In general, there are many ways to
circumvent this limit, such as using predefined \citep{Vaswani2017-transformer}
or relative \citep{dai2019-transformer-xl} sequential embeddings.

\subsection{Modification of input embeddings}
\label{sec:method}

We replace the input embeddings defined in~\eqref{eq:bert-emb} with
\begin{equation}
  \label{eq:2d-emb}
  x_i = s_i + p_{i} + L(\ell_{i}).
\end{equation}
Here, \(\ell_{i}\in\RR^{k}\) stands for \emph{layout embeddings}, which are
described in detail in the next subsection. They carry the information about
the position of the \(i\)-th token on the page.

The dimension \(k\) of the layout embeddings is allowed to differ from the input
embedding dimension \(n\), and this difference is dealt with by a trainable
linear layer \(L\colon \RR^{k}\to\RR^{n}\). However, our main motivation to
introduce the adapter layer \(L\) was to gently increase the strength of the
signal of layout embeddings during training. In this way, we initially avoided
presenting the model with inputs that it was not prepared to deal with. Moreover, in
theory, in the case of non-trainable layout embeddings, the adapter layer may be
able to learn to project \(\ell_{i}\) onto a subspace of the embedding space
that reduces interference with the other terms in \eqref{eq:2d-emb}. For
instance, it is possible for the image of the adapter layer to learn to be
approximately orthogonal to the sum of the remaining terms. This would minimize
any information loss caused by adding multiple vectors. While this was our
theoretical motivation, and it would be interesting to investigate in detail how
much of it actually holds, such detailed considerations of a single model
component exceed the scope of this paper. We included the impact of using the
adapter layer in the ablation study.

We initialize the weight matrix of \(L\) according to a normal distribution
\(\mathcal{N}(0,\sigma^2)\), with the standard deviation \(\sigma\) being a
hyperparameter. We have to choose \(\sigma\) carefully, so that in the initial
phase of training, the \(L(\ell_i)\) term does not interfere overly with the
already learned representations. We experimentally determined the value
\(\sigma=0.02\) to be near-optimal\footnote{we tested the values 0.5, 0.1, 0.02,
  0.004, and 0.0008}.

\subsection{Layout embeddings}
\label{sec:context-embeddings}

In our setting, a document is represented by a sequence of tokens \(t_i\) and
their bounding boxes~\(b_i\). To each element of this sequence, we assign its
layout embedding \(\ell_i\), carrying the information about the position of the
token with respect to the whole document. This could be performed in various
ways. What they all have in common is that the embeddings \(\ell_i\) depend only
on the bounding boxes \(b_i\) and not on the tokens~\(t_i\).

We base our layout embeddings on the method originally used in
\citep{gehring2017convolutional}, and then in \citep{Vaswani2017-transformer} to
define the sequential embeddings. We first normalize the bounding boxes by
translating them so that the upper left corner is at \((0,0)\), and dividing
their dimensions by the page height. This causes the page bounding box to become
\((0, 0, w, 1)\), where \(w\) is the normalized width.

The layout embedding of a token will be defined as the concatenation of four
embeddings of the individual coordinates of its bounding box. For an integer
\(d\) and a vector of scaling factors \(\theta\in\RR^{d}\), we define the
corresponding embedding of a single coordinate \(t\) as
\begin{equation}
  \operatorname{emb_{\theta}}(t) = (\sin(t\theta); \cos(t\theta)) \in \RR^{2d},
\end{equation}
where the \(\sin\) and \(\cos\) are performed element-wise, yielding two vectors
in \(\RR^{d}\). The resulting concatenation of single bounding box coordinate
embeddings is then a vector in \(\RR^{8d}\).

In \citep[Section~3.5]{Vaswani2017-transformer}, and subsequently in other
Transformer-based models with precomputed sequential embeddings, the sequential
embeddings were defined by \(\operatorname{emb}_{\theta}\) with \(\theta\) being
a geometric progression interpolating between \(1\) and \(10^{-4}\). Unlike the
sequential position, which is a potentially large integer, bounding box
coordinates are normalized to the interval \([0,1]\). Hence, for our layout
embeddings we use larger scaling factors \((\theta_r)\), namely a geometric
sequence of length \(n/8\) interpolating between \(1\) and \(500\), where \(n\)
is the dimension of the input embeddings.

\subsection{Relative bias}
\label{sec:relative-bias}

Let us recall that in a typical Transformer encoder, a single attention head
transforms its input vectors into three sequences: queries \(q_{i}\in \RR^{d}\),
keys \(k_{i}\in\RR^{d}\), and values \(v_{i}\in\RR^{d}\). The raw attention
scores are then computed as \(\alpha_{ij} = d^{-1/2}q_{i}^{T}k_{j}\).
Afterwards, they are normalized using softmax, and used as weights in linear
combinations of value vectors.

The point of relative bias is to modify the computation of the raw attention
scores by introducing a bias term: \(\alpha'_{ij} = \alpha_{ij} + \beta_{ij}\).
In the sequential setting, \(\beta_{ij} = W(i-j)\) is a trainable weight,
depending on the relative sequential position of tokens \(i\) and \(j\). This
form of attention bias was introduced in \citep{Raffel2020}, and we will refer
to it as \emph{sequential attention bias}.

We introduce a simple and natural extension of this mechanism to the
two-dimensional context. In our case, the bias \(\beta_{ij}\) depends on the
relative positions of the tokens. More precisely, let \(C \gg 1\) be an integer
resolution factor (the number of cells in a grid used to discretize the
normalized coordinates). If \(b_{i} = (x_{1}, y_{1}, x_{2}, y_{2})\) is the
normalized bounding box of the \(i\)-th token, we first reduce it to a
\(2\)-dimensional position \((\xi_{i},\eta_{i}) = (Cx_{1}, C(y_{1}+y_{2})/2) \),
and then define
\begin{equation}
  \beta_{ij} = H(\lfloor {\xi_{i} - \xi_{j}}  \rfloor) +
  V(\lfloor{\eta_{i}-\eta_{j}} \rfloor),
\end{equation}
where \(H(\ell)\) and \(V(\ell)\) are trainable weights defined for every
integer \(\ell \in [-C, C)\). A good value for \(C\) should allow for a distinction
between consecutive lines and tokens, without unnecessarily affecting
performance. For a typical document \(C=100\) is enough, and we fix this in our
experiments.

This form of attention bias will be referred to as \emph{2D attention bias}. We
suspect that it should help in analyzing, say, tables by allowing the learning
of relationships between cells.


%% file: experiments.tex
All experiments were performed on 8 NVIDIA Tesla V100 32GB GPUs. As our
pretrained base model we used RoBERTa in its smaller, base variant (125M
parameters, 12 layers, 12 attention heads, hidden dimension 768). This was also
employed as the baseline, after additional training on the same dataset we used
for LAMBERT. The implementation and pretrained weights from the
\texttt{transformers} library \citep{pytorch-transformers} were used.

In the LAMBERT model, we used the layout embeddings of dimension \(k=128\), and
initialized the adapter layer \(L\) with standard deviation \(\sigma=0.02\), as
noted in Section~\ref{sec:proposed-method}. For comparison, in our experiments,
we also included the published version of the LayoutLM model
\citep{xu2019layoutlm}, which is of a similar size.

The models were trained on a masked language modeling objective extended with
layout information (with the same settings as the original RoBERTa
\citep{liu2019-roberta}); and subsequently, on downstream information extraction
tasks. In the remainder of the paper, these two stages will be referred to as,
respectively, \emph{training} and \emph{fine-tuning}.

Training was performed on a collection of PDFs extracted from \emph{Common Crawl}
made up of a variety of documents (we randomly selected up to 10 documents from
any single domain). The documents were processed with an OCR system,
\texttt{Tesseract 4.1.1-rc1-7-gb36c}, to obtain token bounding boxes. The final
model was trained on the subset of the corpus consisting of business documents
with non-trivial layout, filtered by an SVM binary classifier, totaling to
approximately 315k documents (3.12M pages). The SVM model was trained on 700
manually annotated PDF files to distinguish between business (e.g.\ invoices,
forms) and non-business documents (e.g.\ poems, scientific texts).

In the training phase, we used the Adam optimizer with the weight decay fix
from~\citet{pytorch-transformers}. We employed a learning rate scheduling method
similar to the one used \bycite{devlin2019-bert}, increasing the learning rate
linearly from \(0\) to \(1\mathrm{e}{-4}\) for the warm-up period of \(10\%\) of
the training time and then decreasing it linearly to \(0\). The final model was
trained with batch size of \(128\) sequences (amounting to \(64\)K tokens) for
approximately \(1000\)k steps (corresponding to training on 3M pages for 25
epochs). This took about 5 days to complete a single experiment.

After training our models, we fine-tuned and evaluated them independently on multiple
downstream end-to-end information extraction tasks. Each evaluation dataset was
split into training, validation and test subsets. The models were extended with a simple classification head on top, consisting of
a single linear layer, and fine-tuned on the task of classifying entity types of
tokens. We employed early stopping based on the \(F_{1}\)-score achieved on the
validation part of the dataset. We used the Adam optimizer again, but this time
without the learning rate warm-up, as it turned out to have no impact on the
results.

The extended model operates as a tagger on the token level, allowing for the
classification of separate tokens, while the datasets contain only the values of
properties that we are supposed to extract from the documents. Therefore, the
further processing of output is required. To this end, we use the pipeline
described in \citep{graliski2020kleister}.

Every contiguous sequence of tokens tagged as a given entity type is treated as
a recognized entity and assigned a~score equal to the geometric mean of the
scores of its constituent tokens. Then, every recognized entity undergoes
a normalization procedure specific to its general data type (e.g.\ date,
monetary amount, address, etc.). This is performed using regular expressions:
for instance, the date \texttt{July, 15th 2013} is converted to
\texttt{2013-07-15}. Afterwards, duplicates are aggregated by summing their
scores, leading to a preference for entities detected multiple times.
Finally, the highest-scoring normalized entity is selected as the output of the
information extraction system. The predictions obtained this way are compared
with target values provided in the dataset using \(F_{1}\)-score as the
evaluation metric. See \citep{graliski2020kleister} for more details.


%% file: results.tex
We evaluated our models on four public datasets containing visually rich
documents. The Kleister NDA and Kleister Charity datasets are part of a larger
Kleister dataset, recently made public \bycite{graliski2020kleister} (many
examples of documents, and detailed descriptions of extraction tasks can be
found therein). The NDA set consists of legal agreements, whose layout variety
is limited. It should probably be treated as a plain-text dataset. The Charity
dataset on the other hand contains reports of UK charity organizations, which
include various tables, diagrams and other graphic elements, interspersed with
text passages. All Kleister datasets come with predefined train/dev/test splits,
with 254/83/203 documents for NDA and 1729/440/609 for Charity.

\begin{table*}[htb]
  \centering
  \input{main-results.tex}
  \label{tab:results}
\end{table*}

The SROIE \citep{icdar-sroie} and CORD \citep{park2019cord} datasets are
composed of scanned and OCRed receipts. Documents in SROIE are annotated with
four target entities to be extracted, while in CORD there are 30 different
entities. We use the public 1000 samples from the CORD dataset with the
train/dev/test split proposed by the authors of the dataset
(respectively, 800/100/100). As for SROIE, it consists of a public training part, and test
part with unknown annotations. For the purpose of ablation studies, we further
subdivided the public part of SROIE into training and test subsets (546/80
documents; due to the lack of a validation set in this split, we fine-tuned for 15
epochs instead of employing early stopping). We refer to this split as SROIE*,
while the name SROIE is reserved for the original SROIE dataset, where the final
evaluation on the test set is performed through the leaderboard
\citep{icdar-sroie-leaderboard}.

In Table~\ref{tab:results}, we present the evaluation results achieved on
downstream tasks by the trained models. With the exception of the Kleister Charity
dataset, where only 5 runs were made, each of the remaining experiments were
repeated 20 times, and the mean result was reported. We compare LAMBERT with
baseline RoBERTa (trained on our dataset) and the original RoBERTa
\cite{liu2019-roberta} (without additional training); LayoutLM
\cite{xu2019layoutlm}; and LayoutLMv2 \cite{xu2020layoutlmv2}. The LayoutLM
model published by its authors was plugged into the same pipeline that we used
for LAMBERT and RoBERTa. In the first four columns we present averaged results
of our experiments, and for CORD and SROIE we additionally provide the results
reported by the authors of LayoutLM, and presented on the leaderboard
\cite{icdar-sroie-leaderboard}.

Since the LayoutLMv2 model was not publicly available at the time of preparing
this article, we could not perform experiments ourselves. As a result some of
the results are missing. For CORD, we present the scores given in
\cite{xu2020layoutlmv2}, where the authors did not mention, though, whether they
averaged over multiple runs, or used just a single model. A similar situation occurs
for LayoutLM; we presented the average results of 20 runs (best run of LAMBERT
attained the score of 95.12), which are lower than the scores presented in
\cite{xu2020layoutlmv2}. The difference could be attributed to using a different
end-to-end evaluation pipeline, or averaging (if the results in
\cite{xu2020layoutlmv2,xu2019layoutlm} come from a single run).

For the full SROIE dataset, most of the results were retrieved from the public
leaderboard \cite{icdar-sroie-leaderboard}, and therefore they come from a
single model. For the base variants of LayoutLM and LayoutLMv2, the results were
unavailable, and we present the scores from the corresponding papers.

In our experiments, the base variant of LAMBERT achieved top scores for all
datasets. However, in the case of CORD, the result reported in
\cite{xu2020layoutlmv2} for the large variant of LayoutLMv2 is superior. If we
consider the best scores of LAMBERT (95.12) instead of the average, and the
scores of LayoutLM reported in \cite{xu2019layoutlm}, LAMBERT slightly
outperforms LayoutLM, while still being inferior to LayoutLMv2. Due to the lack
of details on the results of LayoutLM, it is unknown which of these comparisons
is valid.

For Kleister datasets, the base variant (and in the case of Charity, also the
large variant) of LayoutLM did not outperform the baseline RoBERTa. We suspect
that this might be the result of LayoutLM being better attuned to the evaluation
pipeline used by its authors, and the fact that it was based on an uncased
language model. In the Kleister dataset, meanwhile, performance for entities
such as names may depend on casing.


%% file: main-results.tex
\sisetup{
  table-format=2.2,
  table-number-alignment = right,
}

\caption{Comparison of \(F_{1}\)-scores for the considered models. Best results
  in each column are indicated in bold. In parentheses, the length of training of
  our models, expressed in non-unique pages, is presented for comparison. For
  RoBERTa, the first row corresponds to the original pretrained model without
  any further training, while in the second row the model was trained on our
  dataset. \textsuperscript{a}result obtained from relevant publication;
  \textsuperscript{b}result of a single model, obtained from the SROIE
  leaderboard~\citep{icdar-sroie-leaderboard}}
  
\setlength{\tabcolsep}{4pt}

\begin{tabular}{ cc SSSSSS }

  \toprule
  \multirow{2}{*}{Model} &  \multirow{2}{*}{Params} & \multicolumn{4}{c}{Our experiments} & \multicolumn{2}{c}{External results} \\
  \cmidrule(lr){3-6}\cmidrule(l){7-8}
                         & & {NDA} & {Charity} & {SROIE*} & {CORD} & {SROIE} & {CORD}\\
  \midrule
  RoBERTa \citep{liu2019-roberta}  & 125M & 77.91 & 76.36 & 94.05 & 91.57 & 92.39\textsuperscript{b} & {---}\\
  RoBERTa (16M)                     & 125M & 78.50 & 77.88 & 94.28 & 91.98 & 93.03\textsuperscript{b} & {---}\\[6pt]
  \multirow{2}{*}{LayoutLM \citep{xu2019layoutlm}} & 113M & 77.50 & 77.20 & 94.00 & 93.82 & 94.38\textsuperscript{a} & 94.72\textsuperscript{a}  \\
                         & 343M & 79.14 & 77.13  & 96.48 & 93.62 & 97.09\textsuperscript{b} & 94.93\textsuperscript{a}  \\[6pt]
  \multirow{2}{*}{LayoutLMv2 \citep{xu2020layoutlmv2}} & 200M & {---} & {---}  & {---} & {---} & 96.25\textsuperscript{a} & 94.95\textsuperscript{a} \\
                         & 426M & {---} & {---}  & {---} & {---} & 97.81\textsuperscript{b} & \bf 96.01\textsuperscript{\normalfont a}\\[6pt]
  LAMBERT (16M) & 125M & 80.31 & 79.94 & 96.24 & 93.75 & {---} & {---} \\
  {LAMBERT (75M)} & 125M & \bf 80.42 & \bf 81.34  & \bf 96.93 & \bf 94.41 & \bf 98.17\textsuperscript{\normalfont b} & {---} \\
  \bottomrule \\
\end{tabular}


%% file: ablation.tex
In order to investigate the impact of our modifications to RoBERTa, we performed
an extensive study of hyperparameters and the various components of the final
model. We investigated the dimension of layout embeddings, the impact of the
adapter layer \(L\), the size of training dataset, and finally we performed a
detailed ablation study of the embeddings and attention biases we had used to
augment the baseline model.

In the studies, every model was fine-tuned and evaluated \(20\) times on each
dataset, except for Kleister Charity dataset, on which we fine-tuned every model
\(5\) times: evaluations took much longer on Kleister Charity. For each model
and dataset combination, the mean score was reported, together with the
two-sided 95\% confidence interval, computed using the corresponding
\(t\)-value. We considered differences to be significant when the corresponding
intervals were disjoint. All the results are presented in
Table~\ref{tab:ablation}, which is divided into sections corresponding to
different studies. The \(F_{1}\)-scores are reported as \emph{increases} with
respect to the reported mean baseline score, to improve readability.

\begin{table}[thb]
  \centering
  \input{ablation-table}
  \label{tab:ablation}
\end{table}

\subsection{Baseline}

As a baseline for the studies we use the publicly available pretrained base
variant of the RoBERTa model with 12 layers, 12 attention heads, and hidden
dimension 768. We additionally trained this model on our training set, and
fine-tuned it on the evaluation datasets in a manner analogous to LAMBERT.

\subsection{Embeddings and biases}

In this study we disabled various combinations of input embeddings and attention
biases. The models were trained on 2M pages for 8 epochs, with 128-dimensional
layout embeddings (if enabled). The resulting models were divided into three
groups. The first one contains sequential-only combinations which do not employ
the layout information at all, including the baseline. The second group consists
of models using only the bounding box coordinates, with no access to sequential
token positions. Finally, the models in the third group use both sequential and
layout inputs. In this group we did not disable the sequential embeddings. It
includes the full LAMBERT model, with all embeddings and attention biases
enabled.

Generally, we observe that none of the modifications has led to a
significant performance deterioration. Among the models considered, the only
one which reported a significant improvement for all four datasets---and at the
same time, the best improvement---was the full LAMBERT.

For the Kleister datasets the variance in results was relatively higher than in
the case of SROIE* and CORD. This led to wider confidence intervals, and reduced
the number of significant outcomes. This is true especially for the Kleister NDA
dataset, which is the smallest one. In Kleister NDA, significant improvements
were achieved for both sequential-only models, and for full LAMBERT. The
differences between these increases were insignificant. It would seem that, for
sequential-only models, the sequential attention bias is responsible for the
improvement. But after adding the layout inputs, it no longer leads to
improvements when unaccompanied by other modifications. Still, achieving better
results on sequential-only inputs may be related to the plain text nature of the
Kleister NDA dataset.

While other models did not report significant improvement over the baseline,
there are still some differences between them to be observed. The model using
only 2D attention bias is significantly inferior to most of the others. This
seems to agree with the intuition that relative 2D positions are the least
suitable way to pass positional information about plain text.

In the case of the Kleister Charity dataset, significant improvements were
achieved by all layout-only models, and all models using the 2D attention bias.
Best improvement was attained by full LAMBERT, and two layout-only models using
the layout embeddings; the 2D attention bias used alone improved the results
significantly, but did not reach the top score. The confidence intervals are too
wide to offer further conclusions, and many more experiments will be needed to
increase the significance of the results.

For the SROIE* dataset, except for two models augmented only with a single
attention bias, all improvements proved significant. Moreover, the differences
between all the models using layout inputs are insignificant. We may conclude
that passing bounding box coordinates in any way, except through 2D attention
bias, significantly improves the results. As to the lack of significant
improvements for 2D attention bias, we hypothesize that this is due to its
relative nature. In all other models the absolute position of tokens is somehow
known, either through the layout embeddings, or the sequential position. When a
human reads a receipt, the absolute position is one of the main features used to
locate the typical positions of entities.

For CORD, which is the more complex of the two receipt datasets, significant
improvements were observed only for combined sequential and layout models. In
this group, the model using both sequential and layout embeddings, augmented
with sequential attention bias, did not yield a significant improvement. There
were no significant differences among the remaining models in the group.
Contrary to the case of SROIE*, none of the layout-only models achieved
significant improvement.

\subsection{Layout embedding dimension}

In this study we evaluated four models, using both sequential and layout
embeddings, varying the dimension of the latter. We considered 128-, 384-, and
768-dimensional embeddings. Since this is the same as for the input embeddings
of RoBERTa\textsubscript{BASE}, it was possible to remove the adapter layer
\(L\), and treat this as another variant, in Table \ref{tab:ablation} denoted as
768\textsuperscript{b}.

In Kleister NDA, there were no significant differences between any of the
evaluated models, and no improvements over the baseline. On the other hand, in
Kleister Charity, disabling the adapter layer and using the 768-dimensional
layout embeddings led to significantly better performance. These results remain
consistent with earlier observations that in Kleister NDA the best results were
achieved by sequential-only models, while in the case of Kleister Charity, by
layout-only models. It seems that in the case of NDA the performance is
influenced mostly by the sequential features, while in the case of Charity,
removing the adapter layer increases the strength of the signal of the layout
embeddings, carrying the layout features which are the main factor affecting
performance.

In SROIE* and CORD all results were comparable, with one exception, namely in
SROIE*, the model with the disabled adapter layer did not, unlike the remaining
models, perform significantly better than the baseline.

\subsection{Training dataset size}

In this study, following the observations from \citep{Gururangan2020DontSP}, we
considered models trained on 3 different datasets. The first model was trained
for 8 epochs on 2M unfiltered (see Section~\ref{sec:experiments} for more
details of the filtering procedure) pages. In the second model, we used the same
training time and dataset size, but this time only filtered pages were used.
Finally, the third model was trained for 25 epochs on 3M filtered pages.

It is not surprising that increasing the training time and dataset size, leads
to an improvement in results, at least up to a certain point. In the case of
Kleister NDA dataset, there were no significant differences in the results. For
Kleister Charity, the best result was achieved for the largest training dataset,
with 75M filtered pages. This result was also significantly better than the
outcomes for the smaller dataset. In the case of SROIE* the two models trained
on datasets with filtered documents achieved a significantly higher score than
the one trained on unfiltered documents. There was, in fact, no significant
difference between these two models. This supports the hypothesis that, in this
case, filtering could be the more important factor. Finally, for CORD the
situation is similar to Kleister Charity.


%% file: ablation-table.tex

\caption{Improvements of \(F_{1}\)-score over the baseline for various variants
  of LAMBERT model. The first row (with grayed background) contains the
  \(F_{1}\)-scores of the baseline RoBERTa model. The other grayed row
  corresponds to full LAMBERT. 
  Statistically insignificant improvements over the baseline are grayed. In each
  of three studies, the best result together with all results insignificantly
  smaller are in bold. \textsuperscript{a}filtered datasets;
  \textsuperscript{b}model with a disabled adapter layer }

\setlength{\tabcolsep}{6pt}

\scriptsize

\begin{tabular}{ccccccrrrr}

  \toprule

   \multirow{2}{*}{\rotatebox[origin=c]{90}{\parbox{17mm}{\centering Train epochs and pages}}}
  & \multirow{2}{*}{\rotatebox[origin=c]{90}{\parbox{17mm}{\centering Embeddings dimension}}}
  & \multicolumn{4}{c}{Inputs} & \multicolumn{4}{c}{Datasets} \\

  \cmidrule(r){3-6}\cmidrule(l){7-10}

  &
  & \rotatebox[origin=c]{90}{sequential}
  & \rotatebox[origin=c]{90}{seq. bias}
  & \rotatebox[origin=c]{90}{layout}
  & \rotatebox[origin=c]{90}{2D bias}
  & \multicolumn{1}{c}{NDA}
  & \multicolumn{1}{c}{Charity}
  & \multicolumn{1}{c}{SROIE*}
  & \multicolumn{1}{c}{CORD}\\

  \midrule

  \rowcolor{black!5}
  \multirow{12}{*}{8\(\times\)2M} & \multirow{12}{*}{128}  & \textbullet & & & & \(78.50_{\pm 1.16}\)  & \(77.88_{\pm 0.48}\) & \(94.28_{\pm 0.42}\) & \(91.98_{\pm 0.62}\) \\



  &&& \textbullet &&&  \(\mathbf{1.94}_{\pm 0.46}\) & \gray \(-0.82_{\pm 0.74}\) & \gray \(0.33_{\pm 0.22}\) & \gray \(-0.15_{\pm 0.49}\)\\
  && \textbullet & \textbullet &&& \(\mathbf{2.42}_{\pm 0.61}\) & \gray \(0.52_{\pm 0.64}\) & \(0.79_{\pm 0.17}\) &  \gray \(0.03_{\pm 0.57}\) \\

  \cmidrule{3-10}

  &&&& \textbullet && \gray \(1.25_{\pm 0.59}\) & \(\mathbf{2.62}_{\pm 0.80}\) & \(\mathbf{1.86}_{\pm 0.15}\) & \gray \(0.89_{\pm 0.83}\)\\
  &&&&& \textbullet & \gray \(-0.49_{\pm 0.62}\) & \(2.02_{\pm 0.48}\) & \gray \(0.53_{\pm 0.28}\) & \gray \(-0.17_{\pm 0.62}\) \\
  &&&& \textbullet & \textbullet & \gray \(0.88_{\pm 0.50}\) & \(\mathbf{3.00}_{\pm 0.37}\) & \(\mathbf{1.94}_{\pm 0.16}\) & \gray \({0.68}_{\pm 0.62}\)\\

  \cmidrule{3-10}

  && \textbullet && \textbullet &&  \gray \(1.74_{\pm 0.67}\) & \gray \(0.06_{\pm 0.93}\) & \(\mathbf{1.94}_{\pm 0.18}\) & \(\mathbf{1.42}_{\pm 0.53}\)\\
  && \textbullet &&& \textbullet &  \gray \(1.73_{\pm 0.60}\) & \(2.02_{\pm 0.53}\) & \(\mathbf{2.09}_{\pm 0.22}\) & \(\mathbf{1.93}_{\pm 0.71}\)\\
  && \textbullet && \textbullet & \textbullet & \gray \(0.54_{\pm 0.85}\) & \(1.84_{\pm 0.42}\) & \bf \(\mathbf{2.08}_{\pm 0.38}\) &  \(\mathbf{2.15}_{\pm 0.65}\)\\
  && \textbullet & \textbullet & \textbullet & & \gray \(1.66_{\pm 0.76}\) & \gray \(0.32_{\pm 1.35}\) & \bf \(\mathbf{1.75}_{\pm 0.35}\) & \gray \(1.06_{\pm 0.54}\)\\
  && \textbullet & \textbullet && \textbullet & \gray \(0.85_{\pm 0.91}\) & \(1.84_{\pm 0.27}\) & \bf \(\mathbf{2.01}_{\pm 0.24}\) & \(\mathbf{1.95}_{\pm 0.46}\)\\
  \rowcolor{black!5}
  && \textbullet & \textbullet & \textbullet & \textbullet & \bf \(\mathbf{1.81}_{\pm 0.60}\) & \bf \(\mathbf{2.06}_{\pm 0.69}\) & \(\mathbf{1.96}_{\pm 0.16}\) & \(\mathbf{1.77}_{\pm 0.46}\)\\

  \midrule

  \multirow{4}{*}{8\(\times\)2M} & 128
  & \textbullet && \textbullet && \gray \(1.74_{\pm 0.67}\) & \gray \(0.06_{\pm 0.93}\) & \bf \(\mathbf{1.94}_{\pm 0.18}\) & \(\mathbf{1.42}_{\pm 0.53}\) \\
  & 384 & \textbullet && \textbullet && \gray \(0.90_{\pm 0.54}\) & \gray \(0.70_{\pm 0.40}\) & \bf \(\mathbf{1.86}_{\pm 0.22}\) &  \(\mathbf{1.51}_{\pm 0.60}\) \\
  & 768 & \textbullet && \textbullet && \gray \(0.71_{\pm 1.04}\) & \gray \(0.50_{\pm 0.85}\) & \bf \(\mathbf{2.18}_{\pm 0.25}\) & \(\mathbf{1.54}_{\pm 0.51}\) \\
  & 768\textsuperscript{b} & \textbullet && \textbullet && \gray \(0.77_{\pm 0.58}\) & \bf \(\mathbf{2.30}_{\pm 0.20}\) & \gray \(0.37_{\pm 0.15}\) & \(\mathbf{1.58}_{\pm 0.52}\)\\

  \midrule
  8\(\times\)2M & \multirow{3}{*}{128} & \textbullet & \textbullet & \textbullet & \textbullet &  \(\mathbf{1.81}_{\pm 0.60}\) & \(2.06_{\pm 0.26}\) & \(1.96_{\pm 0.18}\) & \(1.77_{\pm 0.46}\)\\
  8\(\times\)2M\textsuperscript{a} &   & \textbullet & \textbullet & \textbullet & \textbullet & \(\mathbf{1.86}_{\pm 0.66}\) & \(1.92_{\pm 0.19}\) & \bf \(\mathbf{2.60}_{\pm 0.18}\) & \({1.59}_{\pm 0.61}\)\\
  25\(\times\)3M\textsuperscript{a} &   & \textbullet & \textbullet & \textbullet & \textbullet & \bf \(\mathbf{1.92}_{\pm 0.50}\) & \bf \(\mathbf{3.46}_{\pm 0.21}\) & \bf \(\mathbf{2.65}_{\pm 0.13}\) & \(\mathbf{2.43}_{\pm 0.19}\) \\
  \bottomrule\\
\end{tabular}


%% file: conclusions.tex
We introduced LAMBERT, a layout-aware language model, producing contextualized
token embeddings for tokens contained in formatted documents. The model can be
trained in an unsupervised manner. For the end user, the only difference with
classical language models is the use of bounding box coordinates as additional
input. No images are needed, which makes this solution particularly simple, and
easy to include in pipelines that are already based on OCR-ed documents.

The LAMBERT model outperforms the baseline RoBERTa on information extraction
from visually rich documents, without sacrificing performance on documents with
a flat layout. This can be clearly seen in the results for the Kleister NDA
dataset. Its base variant with around 125M parameters is also able to compete
with the large variants of LayoutLM (343M parameters) and LayoutLMv2 (426M
parameters), with Kleister and SROIE datasets achieving superior results. In
particular, LAMBERT\textsubscript{BASE} achieved first place on the Key
Information Extraction from the SROIE dataset leaderboard
\cite{icdar-sroie-leaderboard}.

The choice of particular LAMBERT components is supported by an ablation study
including confidence intervals, and is shown to be statistically significant.
Another conclusion from this study is that for visually rich documents the point
where no more improvement is attained by increasing the training set has not yet
been reached. Thus, LAMBERT's performance can still be improved upon by simply
increasing the unsupervised training set. In the future we plan to experiment
with increasing the model size, and training datasets.

Further research is needed to ascertain the impact of the adapter layer \(L\) on
the model performance, as the results of the ablation study were inconclusive.
It would also be interesting to understand whether the mechanism through which
it affects the results is consistent with the hypotheses formulated in
Section~\ref{sec:proposed-method}.

\subsubsection{Acknowledgments.}
\label{sec:acknowledgments}

The authors would like to acknowledge the support the Applica.ai project has
received as being co-financed by the European Regional Development Fund
(POIR.01.01.01--00--0605/19--00).
